\documentclass[runningheads]{llncs}
\newcommand{\ie}{\textit{i.e., }}
\newcommand{\eg}{\textit{e.g., }}
\usepackage{algorithm}
\usepackage{algorithmic}
\usepackage{soul}
\usepackage{url}
\usepackage[hidelinks]{hyperref}
\usepackage[utf8]{inputenc}
\usepackage{graphicx}
\usepackage{amsmath}
\usepackage{booktabs}
\usepackage{algorithm}
\usepackage{algorithmic}
\usepackage{multirow}
\usepackage[normalem]{ulem}
\useunder{\uline}{\ul}{}
\usepackage{subfigure}
\usepackage{color}
\usepackage{hyperref}
\usepackage{amssymb}
\usepackage{colortbl}
\usepackage{caption}
\usepackage[misc]{ifsym}
\def\C#1{\mathcal #1}
\usepackage{bm}

\begin{document}
\title{Dual Graph Multitask Framework for Imbalanced Delivery Time Estimation}
\titlerunning{DGM-DTE}

\author{
Lei Zhang\inst{1,2} \and
Mingliang Wang\inst{3} \and
Xin Zhou\inst{4} \and
Xingyu Wu\inst{3} \and
Yiming Cao\inst{1,2} \and
Yonghui Xu\inst{2} \and
Lizhen Cui\inst{1,2}\Letter \and
Zhiqi Shen\inst{4}\Letter \\
}
\authorrunning{L. Zhang et al.}
\institute{
School of Software, Shandong University, China \and
Joint SDU-NTU Centre for Artificial Intelligence Research (C-FAIR), Shandong University, China \and
Alibaba Group, China \and
School of Computer Science and Engineering, Nanyang Technological University, Singapore\\
\email{clz@sdu.edu.cn, zqshen@ntu.edu.sg}
}

\maketitle              

\begin{abstract}
Delivery Time Estimation (DTE) is a crucial component of the e-commerce supply chain that predicts delivery time based on merchant information, sending address, receiving address, and payment time. Accurate DTE can boost platform revenue and reduce customer complaints and refunds. However, the imbalanced nature of industrial data impedes previous models from reaching satisfactory prediction performance. Although imbalanced regression methods can be applied to the DTE task, we experimentally find that they improve the prediction performance of low-shot data samples at the sacrifice of overall performance. To address the issue, we propose a novel Dual Graph Multitask framework for imbalanced Delivery Time Estimation (DGM-DTE). Our framework first classifies package delivery time as head and tail data. Then, a dual graph-based model is utilized to learn representations of the two categories of data. In particular, DGM-DTE re-weights the embedding of tail data by estimating its kernel density. We fuse two graph-based representations to capture both high- and low-shot data representations. Experiments on real-world Taobao logistics datasets demonstrate the superior performance of DGM-DTE compared to baselines.


\end{abstract}

\section{Introduction}
As e-commerce proliferates, e-commerce logistics becomes a major industry focus, and Delivery Time Estimation (DTE) is an important part of intelligent e-commerce logistics. Accurate DTE can enhance the users' shopping experience and increase the purchase rate to raises platform revenue~\cite{cui2020sooner}.

\begin{figure*}[t]
\centering
\includegraphics[width=0.7\textwidth]{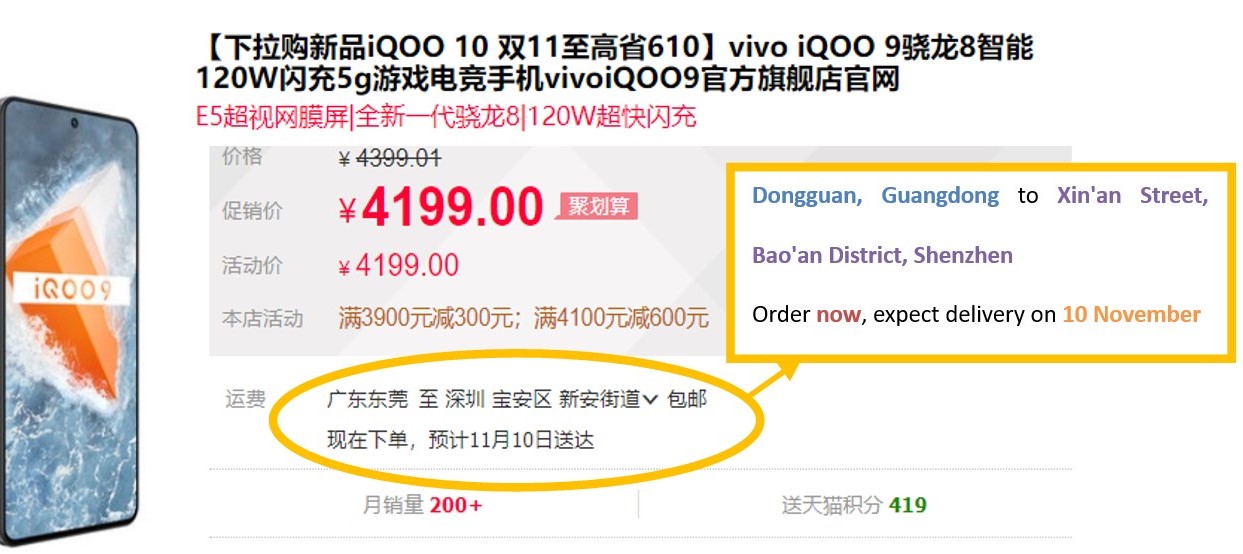}
\caption{A DTE example on Taobao platform.}
\label{fig:example}
\end{figure*}

In industrial e-commerce logistics scenarios, we focus on a category of Origin-Destination (OD) DTE problems, where the delivery time of orders is predicted based on known attributes, such as order merchant, sending address, receiving address, and payment time. Fig.~\ref{fig:example} presents a demonstration example of DTE when the user browses an item on the Taobao e-commerce platform.

\begin{figure}[t]
 \centering
 \subfigure[Order distribution]{
  \begin{minipage}{0.47\columnwidth}
   \centering
   \label{fig:motivation1}
   \includegraphics[width=\textwidth]{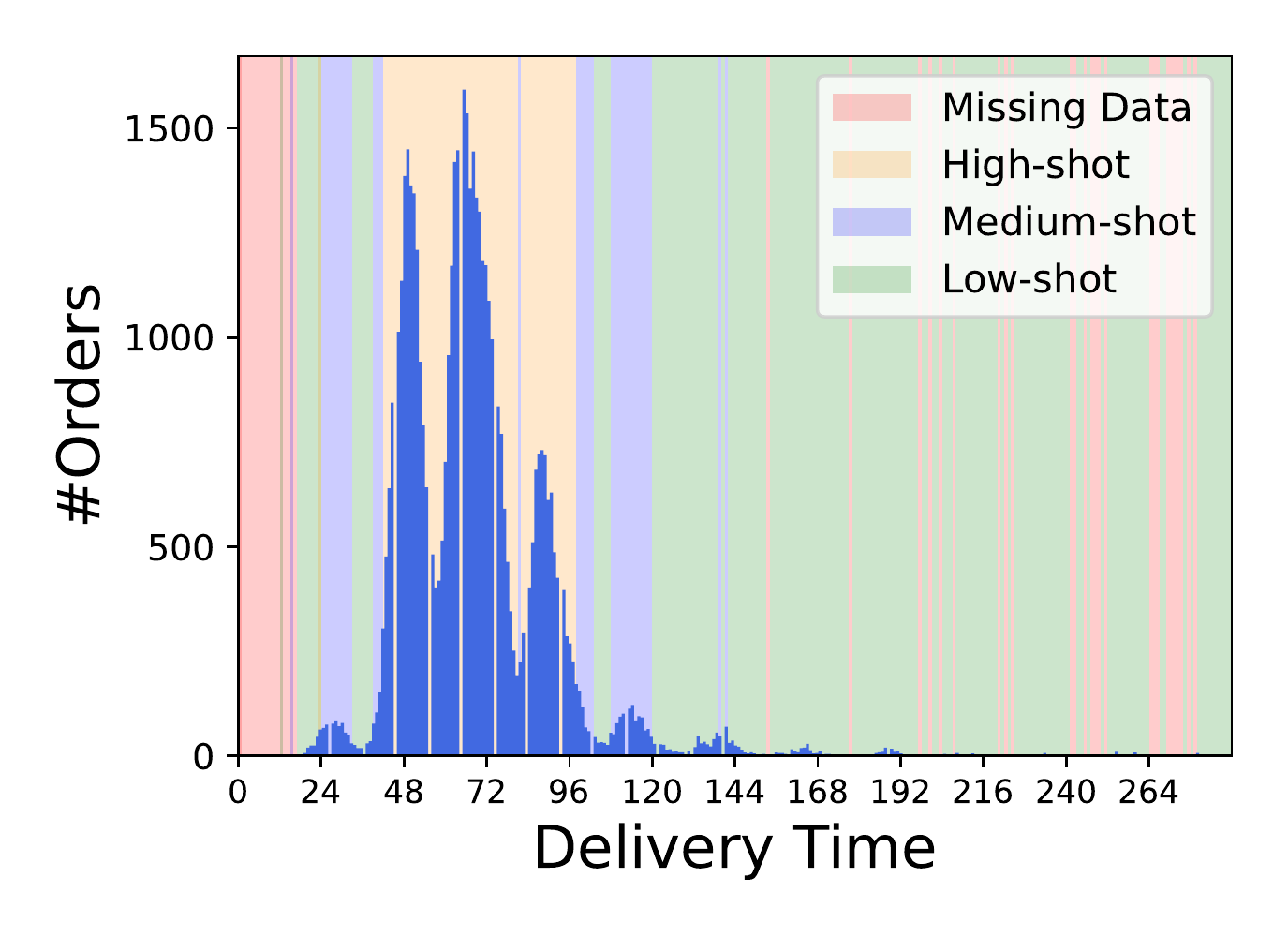}
  \end{minipage}
 }
 \subfigure[MAE distribution with delivery time]{
  \begin{minipage}{0.47\columnwidth}
   \centering
   \label{fig:motivation2}
   \includegraphics[width=\textwidth]{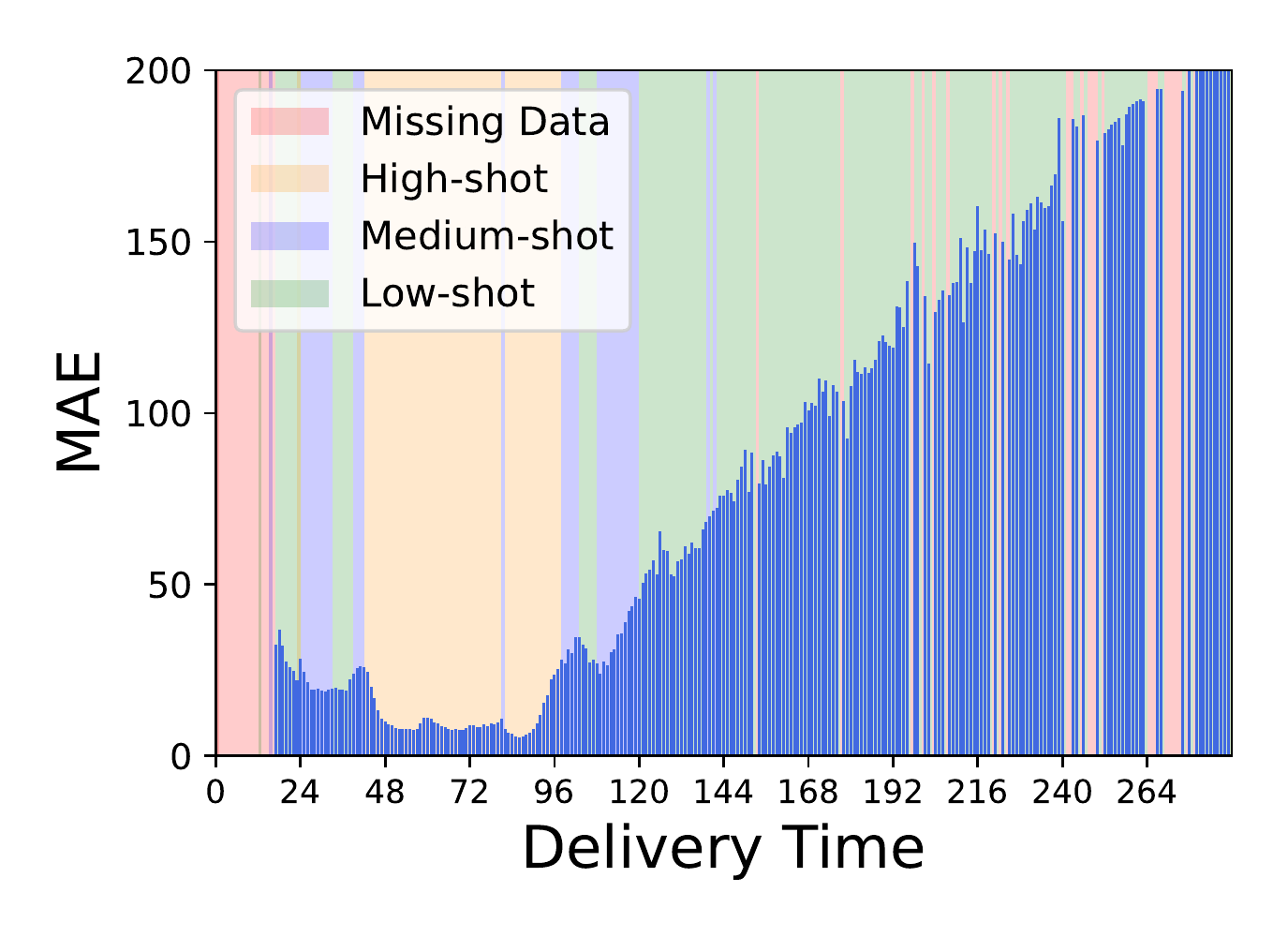}
  \end{minipage}
 }
 \caption{Order distribution and MAE distribution with different delivery time.}
 \label{fig:motivation}
\end{figure}

Existing research formalizes the OD DTE problem as a regression problem, which uses end-to-end models such as Deep Neural Networks (DNNs) and representation learning~\cite{AraujoE21,JindalQCSY17,LiFWSYL18} to predict the delivery time based on the order features. 
However, industrial e-commerce logistics data exhibits a skewed distribution of orders, \ie imbalanced data, as shown in Fig.~\ref{fig:motivation1}. 
Most of the orders (about 90\%) are delivered within 48-96 hours (\ie the high-shot data region), with a portion of the data still in the medium-shot region (6.6\%) and low-shot region (3.3\%). As a result, models trained with such severely imbalanced data may have inferior performance on the medium- and low-shot data, as shown in Fig.~\ref{fig:motivation2}. Besides, we find that the predicted values of these models are typically smaller than the real delivery time for orders within medium- and low-shot regions. 
Consequently, the platform may observe increasing user complaints and refund rates as orders cannot be received within the predicted time.

Dealing with imbalanced data in e-commerce logistics scenarios is a pressing challenge. There are two lines of research on imbalanced regression: synthesizing new samples for rare labeled data~\cite{BrancoTR17} and loss re-weighting~\cite{SteiningerKDKH21,YangZCWK21}. Although these methods improve prediction performance for rare labeled data, they sacrifice prediction and representation performance for high-shot data, as shown in Table~\ref{tab:high-low}. Besides, current performance tests for imbalanced regressions are conducted on balanced test data, which does not make sense in practical industrial applications, where the test data is frequently also imbalanced.

To address the above challenges, this paper proposes a Dual Graph Multitask framework for imbalanced Delivery Time Estimation (DGM-DTE). Specifically, DGM-DTE first performs a classification task, which divides the orders into head and tail data according to the delivery time.
Then, we leverage a dual graph-based representation module, one learns high-shot data representation in head data, and another re-weights the representations of tail data according to kernel density estimation of labels. For graph-based representation module, we build relation graphs from spatial, temporal, and merchant attributes of orders and use graph neural network (GNN) to capture both inter- and intra-correlations of attributes. Besides, we employ a simple but effective normalization for embedding de-biasing. The order representations learned from dual graph module are then aggregated, so that the model can focus on both high-shot regional data and rare labeled data. Overall, we propose a multitask learning framework that predicts delivery time from two-view (classification and imbalanced regression). 

The main contributions of this paper are as follows.

\begin{itemize}
    \item We focus on the imbalanced distribution of industrial e-commerce logistics data and propose a dual graph multitask model for imbalanced delivery time prediction.
    \item We design a GNN-based order representation module that can fully exploit the inter- and intra-correlation of order attributes. 
    \item We conduct extensive experiments on real datasets from the Taobao platform to demonstrate the effectiveness of DGM-DTE in prediction performance. Various ablation studies validate the design of DGM-DTE is capable of improving the prediction accuracy of medium- and low-shot orders without compromising its performance on high-shot orders.
\end{itemize}

\section{Related Work}
\subsection{Delivery Time Estimation}
DTE is a category of estimated time of arrival problems, which is widely studied in transportation~\cite{FanXZL22,HongLYLFWQY20}, logistics~\cite{AraujoE21,LiWWLWDM21,Xinwsdm23}, and food delivery~\cite{GaoZWHRHHS21}. OD method is a line of research that predicts arrival time based on origin and destination without actual trajectories. \cite{WangTKKL19} proposes a simple baseline model, which finds similar trips based on adjacent origin and destination. \cite{BertsimasDJM19} introduces a simple static model with network optimization to predict travel time. \cite{JindalQCSY17,AraujoCA19} directly use DNN for end-to-end prediction. The above methods only use travel features for prediction without considering relations between attributes. MURAT~\cite{LiFWSYL18} utilizes a multitask learning with GNNs to leverage the road network and spatio-temporal priors. Besides, BGE~\cite{LiWWLWDM21} is a bayesian graph model learning observed and unobserved attributes in logistics. However, existing methods ignore the imbalanced nature of data, resulting in unsatisfactory prediction performance.

\subsection{Imbalanced Regression}
The research of imbalanced regression is still in its initial stage, which can be divided into two streams: re-sampling and re-weighting. \cite{BrancoTR17,TorgoRPB13,BrancoTR18} introduce pre-processing strategies to re-sample and synthesize new samples for rare labeled data. Another line of research proposes the re-weight loss function to deal with imbalanced data. DenseWeight~\cite{SteiningerKDKH21} weights the data based on the sparsity of the target value by kernel density estimation. \cite{YangZCWK21} proposes two algorithms to smooth the distribution of labels and features. Besides, \cite{RenZY022} designs a Balanced MSE loss function, which uses the training label distribution prior to recover the balanced prediction. However, existing methods compromise prediction performance over high-shot data and test the performance with balanced data, which is impractical in industrial applications.

\begin{figure*}[t]
\centering
\includegraphics[width=0.8\textwidth]{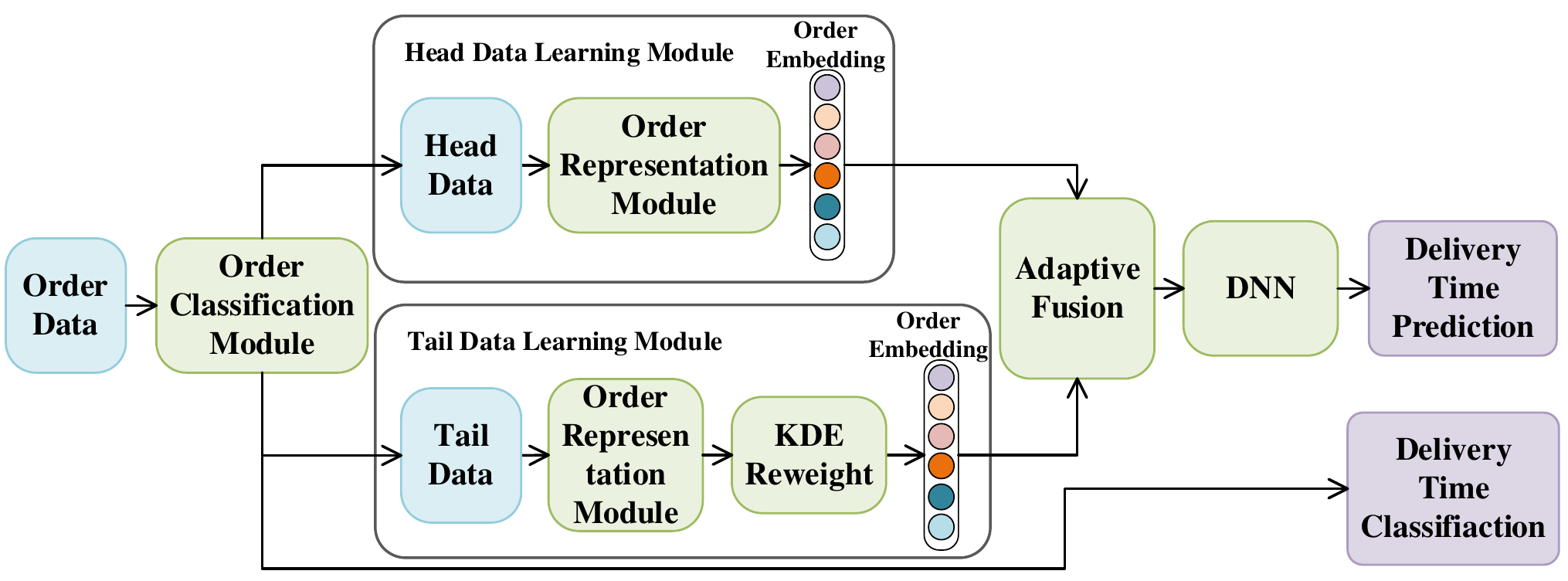}
\caption{The overall DGM-DTE framework.}
\label{fig:model1}
\end{figure*}


\section{The Proposed Model}
The overall framework of the proposed DGM-DTE model is shown in Fig.~\ref{fig:model1}. Firstly, we propose a classification module to divide orders into head and tail data. Then, we use the dual graph-based order representation module to learn the head and tail data embeddings separately. Finally, we aggregate the two parts of data embeddings for delivery time regression prediction.

\begin{figure*}[t]
\centering
\includegraphics[width=0.6\textwidth]{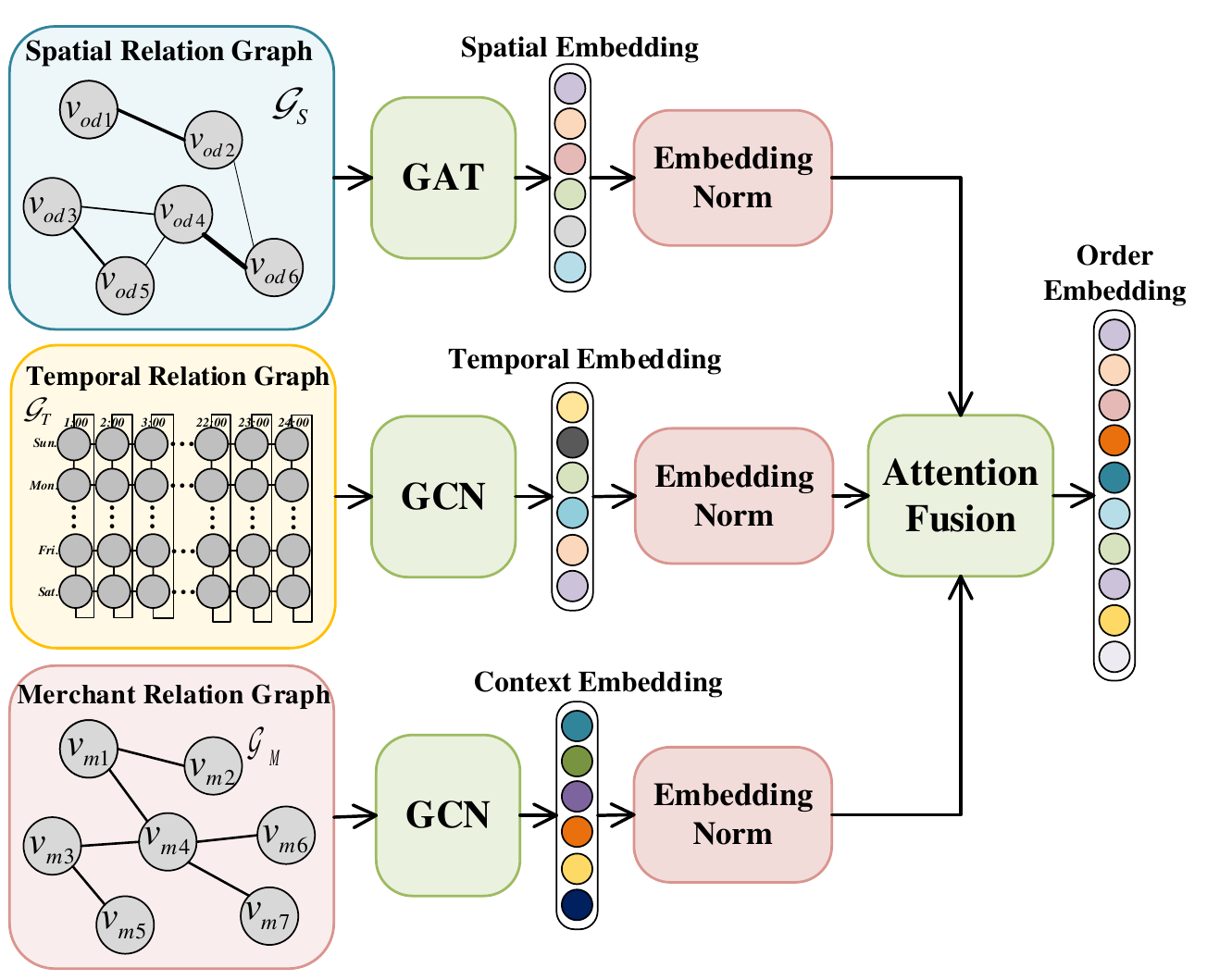}
\caption{Graph-based order representation model.}
\label{fig:model2}
\end{figure*}

\subsection{Graph-based Order Representation Learning}

The graph-based order representation module learns order embedding from the inter- and intra-correlation of order attributes, its structure is shown in Fig.~\ref{fig:model2}. 

We construct three graphs for three main attributes of orders, named spatial, temporal, and merchant relation graphs. For the weighted spatial relation graph $\C{G}_S=(\C{V}_{OD},\C{E}_S)$, where $\C{V}_{OD}$ denotes the node set composed of OD pairs (\ie a group of sending and receiving addresses) of orders; $\C{E}_S$ denotes a set of edges, which represents the relation of OD pairs in term of geographic location. 
For example, we define the weight of two OD pairs as the sum of the distance between their origins and the distance between their destinations.
As for the unweighted temporal graph $\C{G}_T=(\C{V}_{T},\C{E}_T)$, which represents the periodicity of payment time in weeks and days. Each node in the temporal graph denotes the payment timestamp in hours, which is connected to adjacent hour nodes and its counterparts at the same hour of a week. Besides, the merchant graph $\C{G}_M=(\C{V}_{M},\C{E}_M)$ represents the similarity between merchants, which is manually defined based on historical orders.

To fully exploit the inter-correlation of attributes, GNN is used to learn node embeddings of attribute relation graphs. For the weighted spatial graph $\C{G}_S$, we leverage Graph Attention Network (GAT) to learn OD node embedding,
\begin{equation}
    \bm{E}_{OD_i}=\sigma\left(\frac{1}{K} \sum_{k=1}^{K} \sum_{j \in \mathcal{N}\left(\mathcal{V}_{OD_i}\right)} \alpha_{i j} \bm{W} \bm{X}_{OD_j}\right),
\label{eq4}
\end{equation}
where $\bm{X}_{OD_j}\in \mathbb{R}^{F_{OD}}$ denotes the initial feature of node ${\C{V}_{OD}}_{j}$, $K$ denotes the number of head in multi-head attention. $\bm{W}$ is a trainable weight matrix, and $\mathcal{N}\left(\mathcal{V}_{OD_i}\right)$ is the neighbors of node $\C{V}_{OD_i}$; and $\alpha_{i j}$ denotes the attention coefficient, which can be calculated as follows:
\begin{equation}
\alpha_{i j}=\frac{\exp \bigl(\operatorname{ReLU}\left(f_{a}([\bm{W} \bm{X}_{OD_{i}} \mid \bm{W} \bm{X}_{OD_{j}}])\right)\bigr)}{\sum_{k \in \mathcal{N}\left({\mathcal{V}_{OD_i}}\right)} \exp \bigl(\operatorname{ReLU}\left(f_{a}([\bm{W} \bm{X}_{OD_{i}} \mid \bm{W} \bm{X}_{OD_{k}}])\right)\bigr)},
\end{equation}
where $f_{a}(\cdot)$ is a fully connected neural network.
For the unweighted temporal graph $\C{G}_T$, Graph Convolution Network (GCN) is utilized to learn the temporal node embedding.
\begin{equation}
    \bm{E}_{T}^{l}=\operatorname{GCN}\left(\bm{X}_{T}, \bm{A}_{T}\right)=\sigma\left(\widetilde{\bm{D}}_{T}^{-\frac{1}{2}} \widetilde{\bm{A}}_T \widetilde{\bm{D}}_{T}^{-\frac{1}{2}} \bm{{E}}_{T}^{l-1} \bm{W}\right),
\end{equation}
where, $\widetilde{\bm{A}}_T=\bm{A}_T+\bm{I}$, $\widetilde{\bm{D}}$ denotes the degree matrix of $\widetilde{\bm{A}}_T$, $\bm{A}_T$ and $\bm{X}_T \in \mathbb{R}^{N_{T} \times F_T}$ are the adjacency matrix and initial node features of $\C{G}_T$. Similarly, the merchant embedding can be learned via GCN as $\bm{E}_{M}=\operatorname{GCN}\left(\bm{X}_{M}, \bm{A}_{M}\right)$.

As representations of imbalanced data suffer from biased embedding, especially for attribute embedding.
We further propose a simple but effective normalization method for embedding de-biasing on each attribute node embedding. 
Specifically, we use a per-dimension normalization to alleviate embedding bias, which calculates as $\bm{e}_{mn}^{norm}=\frac{\bm{e}_{mn}}{\|\bm{e}_{n}\|}$.

Finally, we use an attention fusion to adaptively learn intra-correlation representation (\ie order embedding) according to spatial, temporal, and merchant embeddings. Due to varying contributions of different attributes to the order representation, we learn aggregation coefficients through a multi-head attention mechanism~\cite{VaswaniSPUJGKP17}. We use $\bm{Q}=\bm{E}_{O D} \bm{W}^{q} , \quad \bm{K}=\bm{E}_{T} \bm{W}^{k}, \quad \bm{V}=\bm{E}_{{M}} \bm{W}^{v}$ as query, key, and value, respectively. $\bm{W}^{q} \in \mathbb{R}^{d_{OD} \times {d}_{O}}$, $\bm{W}^{k} \in \mathbb{R}^{d_T \times {d}_{O}}$, and $\bm{W}^{v} \in \mathbb{R}^{d_m \times {d}_{O}}$ are weight matrices. The order embedding $\bm{E}_{O} \in \mathbb{R}^{N \times {d}_{O}}$ can be represented as,
\begin{equation}
    \begin{array}{c}
\bm{E}_{O}=\operatorname{CONCAT}\left( head_{1}, \ldots, head_{h}\right) \bm{W}^{O} \\
head_{i}= {\operatorname{Attention}}\left(\bm{Q} \bm{W}_{i}^{\bm{Q}}, \bm{K} \bm{W}_{i}^{\bm{K}}, \bm{V} \bm{W}_{i}^{\bm{V}}\right).
\end{array}
\end{equation}

\subsection{Delivery Time Classification}
As Fig.~\ref{fig:motivation} reveals, the imbalance of logistics data is mainly reflected in the large amount of data concentrated in the part of the header region. In contrast, the volume of data in the tail is small and spread over a wide area. Besides, it is also important to predict whether a package will arrive within a certain period (\eg three-day delivery) in practical e-commerce.

Therefore, we classify the delivery time as an auxiliary task of multitask learning. The division of binary classification is based on a defined time $t_c$, where orders with delivery time greater than $t_c$ are in one category, \ie tail orders $z_{i,t}$, and vice versa are head orders $z_{i,h}$. In the order classification module, we first use a graph-based order representation to learn the order embedding $\bm{E}_{O}$, then follow a MLP to obtain the classified output (\ie the prediction probability), $\bm{z}_{i}=MLP(\bm{E}_{O})=[z_{i,h},z_{i,t}]$, and $\widehat{y_c}_{i}=argmax (\bm{z}_{i})$ is the classified prediction.

\subsection{Dual Graph-based Order Representation}

We propose a novel dual graph-based order representation learning for imbalanced regression learning, one for learning high-shot data representation in head data, and another for mining imbalanced tail data representation.

For the head data learning module, the input is the head data $\bm{O}^{head}=\bm{O}(\widehat{y_c}_{i}=0)$ predicted by the classification module, and the output (\ie order embedding) $\bm{E}_{O}^{head}$ is learned from the graph-based order representation module. However, it is inappropriate to directly use the same module for tail data $\bm{O}^{tail}=\bm{O}(\widehat{y_c}_{i}=1)$, as the tail data contains a wide range of delivery time and also shows an imbalanced distribution. So we propose an embedding re-weight strategy for the tail data learning module. A kernel density estimation is used to learn the imbalance property corresponding to continuous targets~\cite{YangZCWK21},
\begin{equation}
    \widetilde{p}\left(\bm{y}_t^{\prime}\right) \triangleq \int_{\bm{\mathcal{Y}}} \mathrm{k}\left(\bm{y}_t, \bm{y}_t^{\prime}\right) p(\bm{y}_t) d y,
\end{equation}
where $\mathrm{k}\left(y_t, y_t^{\prime}\right)$ is the Gaussian kernel function for tail label, and delivery time label are in the label space $\bm{\mathcal{Y}}$, \ie $y_t, y_t^{\prime} \in \bm{\mathcal{Y}}$. Then, we can define the weight by square inverse of label density distribution, $\bm{w}_t=1/\sqrt{ \widetilde{p}\left(\bm{y}_t^{\prime}\right)}$. The order embedding of tail data represented as $\bm{E}_{O}^{tail\prime}=\bm{w}_t \bm{E}_{O}^{tail}$.

\subsection{Adaptive Delivery Time Prediction and Model Training}
We leverage a DNN for regression prediction based on order representations for the main task delivery time prediction. First, an adaptive fusion is used to merge by index the head data embedding and re-weighted tail data embedding, 
The predicted delivery time is
 $\widehat{y}_r=\operatorname{DNN}(merge(\bm{E}_{O}^{head},\bm{E}_{O}^{tail\prime}))$.

Our proposed multitask framework has two tasks: delivery time classification as the auxiliary task and delivery time estimation as the main task. So the loss function contains two parts: Mean Absolute Error (MAE) for regression prediction and binary cross-entropy for classification,
\begin{equation}
        \mathcal{L}=MAE(\bm{Y}_r, \widehat{\bm{Y}}_r)+BCE(\bm{Y}_c, \widehat{\bm{Y}}_c).
\end{equation}

\section{Experiments}

\subsection{Experimental Settings}
\noindent \textbf{Dataset.} We collect two real e-commerce logistics datasets from Taobao, one of the world's largest e-commerce platforms. The first dataset, ``D1", contains 452,917 orders received in Weihai, Shandong, China. The second dataset, named ``D2", contains 1,048,575 orders with receiving addresses in Hangzhou, Zhejiang, China. We use time as the basis for dataset division since our goal is to predict the delivery time of future orders based on historical data. The details and division of the datasets are shown in Table~\ref{tab:datasets}.

\begin{figure}[t]
 \centering
 \subfigure[Order distribution in D1]{
  \begin{minipage}{0.47\columnwidth}
   \centering
   \label{fig:weihai}
   \includegraphics[width=\textwidth]{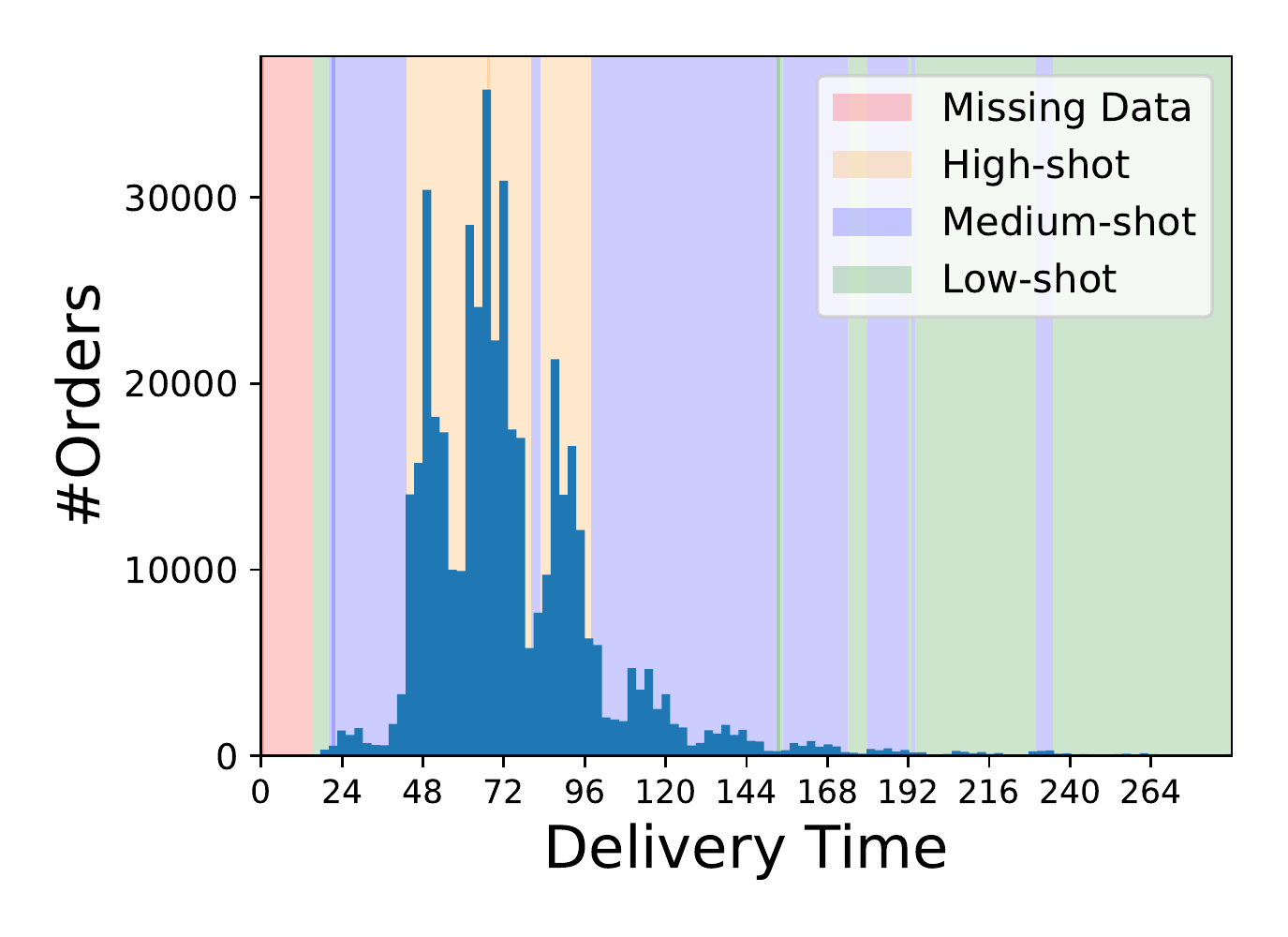}
  \end{minipage}
 }
 \subfigure[Order distribution in D2]{
  \begin{minipage}{0.47\columnwidth}
   \centering
   \label{fig:hangzhou}
   \includegraphics[width=\textwidth]{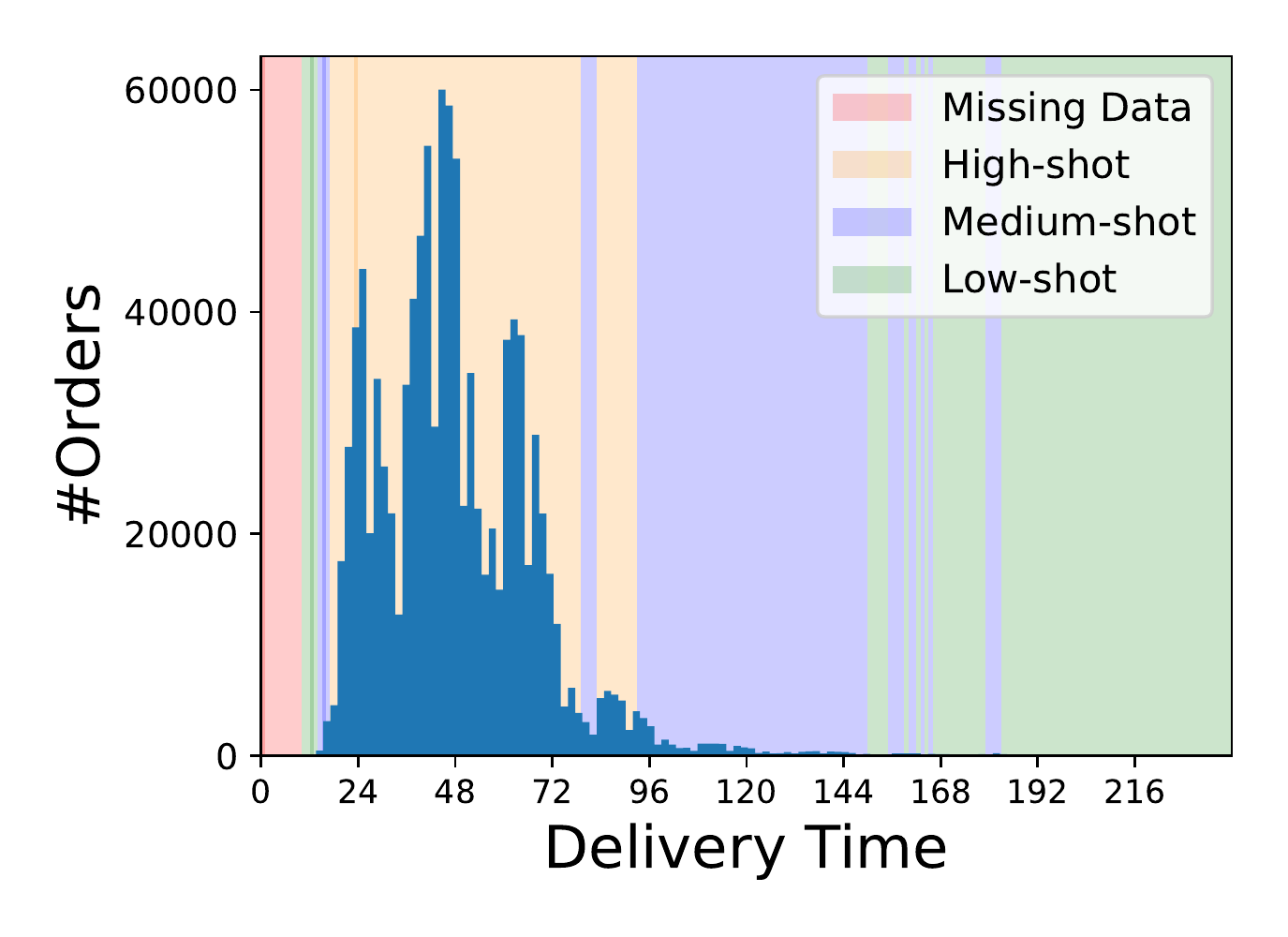}
  \end{minipage}
 }
 \caption{Order distribution in datasets.}
 \label{fig:dis}
\end{figure}

To explore the imbalanced distribution of the dataset, we analyze the order distribution of two datasets, as shown in Fig.~\ref{fig:dis}. Overall, both datasets exhibit seriously imbalanced distributions, with the more significant imbalance in dataset D1 manifesting as a wider distribution and a higher proportion of low-shot data. 
For D1, most orders are delivered within 48-96 hours, but a few orders take an incredibly long time (\textgreater10 days). The delivery time of orders in D2 is generally shorter than D1. Most orders in D2 are delivered within 24-72 hours, and the delivery time for low-shot data is typically greater than one week.

\begin{table}[tbp]
\small
\renewcommand\arraystretch{1.2}
  \centering
  \scriptsize
  \caption{The Statistics of Experimental Datasets.}
    \begin{tabular}{l|c|c|c|c|c|c|c|c|c}
    \toprule
     & \textbf{\#Order}    & \textbf{\#Merchant}    & \textbf{\#Sender} &   \textbf{\#Receiver} & \textbf{\#Day} & \textbf{\#OD pairs} & \begin{tabular}[c]{@{}c@{}}\textbf{\#Train} \\ \textbf{day}\end{tabular}  &\begin{tabular}[c]{@{}c@{}}\textbf{\#Val.} \\ \textbf{day}\end{tabular}  &\begin{tabular}[c]{@{}c@{}}\textbf{\#Test} \\ \textbf{day}\end{tabular} \\
    \hline
    D1 &452,917 &7,636 & 1,004 &9 &110 &3,679 &90 &10 &10 \\
    \hline
    D2 &1,048,575 &9,954 &1,152 &57 &51 &10,689 &37 &7 &7 \\
    \bottomrule
    \end{tabular}%
  \label{tab:datasets}%
\end{table}%

\noindent \textbf{Evaluation Metrics.}
For the DTE task, we use the general regression metrics, \ie MAE, Mean Absolute Percentage Error (MAPE), and Window of Error (EW)~\cite{AraujoE21} to evaluate the prediction performance. Where EW calculated as $p=\frac{1}{N} \sum_{i}^{N} H(EW-|y_{r}-\hat{y}_r|)$, $H(\cdot)$ is the Heaviside step function, and p set as 90\% , which measures the error window for 90\% of orders.

\noindent \textbf{Implementation details.}
Our model is implemented on PyTorch and trained with a batch size of 2048. In the graph-based order representation module, the initial node feature sizes of spatial, temporal, and merchant graphs are 207, 61, and 128, and node embedding dimensions after two layers GNN (\ie GAT, GCN) is 64 for all. The classification thresholds $t_c$ for D1 and D2 datasets are 96 and 72 hours, respectively. The order embedding size is 128, and the neuron numbers of DNN for regression prediction are 128, 64, and 32. We use Adam as the optimizer to train the model, and the learning rate is 0.0005.

\noindent \textbf{Baselines.}
We compare our proposed DGM-DTE model with two types of baseline approaches for performance evaluation. The first type of baseline is the OD DTE models for the comparison of prediction performance, \textit{e.g.}, \textbf{TEMP}~\cite{WangTKKL19} finding similar orders, \textbf{XGBoost}~\cite{ChenG16}, \textbf{xDeepFM}~\cite{LianZZCXS18}, and \textbf{STNN}~\cite{JindalQCSY17} directly using order features as the input, and graph-based model \textbf{MURAT}~\cite{LiFWSYL18} and \textbf{BGE}~\cite{LiWWLWDM21}. 
Another type is the imbalanced models for validating the capability of the prediction for imbalanced data, such as \textbf{LDS}~\cite{YangZCWK21} and \textbf{BMSE}~\cite{RenZY022} for designing the re-weight loss function, and \textbf{SMOGN}~\cite{BrancoTR17} synthesizing data for pre-processing the imbalanced data.

\begin{table}[htbp]
\renewcommand\arraystretch{1}
  \centering
  \footnotesize
  \caption{DTE performance of different methods in terms of MAE, MAPE, and EW. The best results are \textbf{bold faces}, and the second best results are \underline{underlined}.}
    \begin{tabular}{l|c|c|c|c|c|c|c|c|c|c}
    \toprule
          \multicolumn{2}{c|}{} &\multicolumn{3}{c|}{\textbf{D1}} & \multicolumn{3}{c|}{\textbf{D2}} & \multicolumn{3}{c}{\textbf{Balanced Test}} \\
    \hline
        \textbf{Model} &
        \textbf{Types} &
        \multicolumn{1}{c|}{\textbf{MAE}} & \multicolumn{1}{c|}{\textbf{MAPE}} & \multicolumn{1}{c|}{\textbf{EW}} & \multicolumn{1}{c|}{\textbf{MAE}} & \multicolumn{1}{c|}{\textbf{MAPE}} & \multicolumn{1}{c|}{\textbf{EW}} & \multicolumn{1}{c|}{\textbf{MAE}} & \multicolumn{1}{c|}{\textbf{MAPE}} & \multicolumn{1}{c}{\textbf{EW}} \\
    \hline
    TEMP & \multirow{6}{*}{\begin{tabular}[c]{@{}c@{}} OD \\ DTE\end{tabular}} & 17.31 & 24.56\% & 30.05 & 11.22 & 23.56\% & 20.67 & 85.24 & 78.38\% & 178.38  \\
    XGBoost & & 16.15 & 24.43\% & \underline{28.68} & 11.04 & 23.40\% & \underline{20.01} & 88.62 & 60.12\% & 189.97 \\
    STNN & & 17.74 & 27.98\% & 33.73 & 12.57 & 24.46\% & 24.21 &89.84  &62.01\% &186.97  \\
    BGE &    & 15.61 & 25.36\% & 29.73 & \underline{10.26} & 20.46\% & 20.67 & 87.02 & 61.19\% & 182.38 \\
    MURAT & & 17.74 & 29.52\% & 31.30 & 14.81 & 30.67\% & 25.08 &86.28  &59.96\% &188.32  \\
    xDeepFM & & 16.76 & 39.49\% & 33.52 & 13.49 & 37.43\% & 29.64 & 86.11& 69.98\% &181.30  \\
    \hline 
    LDS & \multirow{3}{*}{\begin{tabular}[c]{@{}c@{}} Imba-\\lanced \\ Models\end{tabular}} & 17.15 & 25.65\% & 32.31 & 13.58 & 22.25\% & 22.08 & 86.46 & \underline{57.73\%} & 196.25 \\
    BMSE &  & \underline{13.87} & \underline{19.12\%} & 28.82 & 10.68 & \underline{19.26\%} & 21.02 & \underline{84.88} & 59.16\% & 193.42 \\
    SMOGN &  & 20.75 & 28.91\% & 38.12 & 14.55 & 24.28\% & 24.37 & 85.09 & 60.24\% & \underline{173.77} \\
    \hline
    DGM-DTE & Ours & \textbf{11.97} & \textbf{16.81\%} & \textbf{22.43} & \textbf{8.52}  & \textbf{17.74\%} & \textbf{19.77} & \textbf{83.30}  & \textbf{57.43\%} & \textbf{172.21} \\
    \bottomrule
    \end{tabular}%
  \label{tab:performance}%
\end{table}%

\begin{table*}[t]
\centering
\small
\renewcommand\arraystretch{1.1}
\caption{The MAE performance of different models in term of high-, medium-, and low-shot region data. The best results are in \textbf{bold faces}, and the second best results are \underline{underlined}. The imrpov. is the improvement of ours vs. im-reg.}
\begin{tabular}{c|c|c|c|c|c|c|c|c}
\hline
Shot                  & XGBoost & BGE    & LDS & BMSE  & SMOGN   & im-reg & DGM-DTE & improv. \\ \hline
High &13.5 &12.04 &22.46 &12.18 &35.78 &\underline{9.32} &\textbf{9.23} & 1.0\% \\
\hline
Medium &32.98 &34.55 &\textbf{30.35} &36.28 &32.76 &32.84 &\underline{31.88} & 2.9\% \\
\hline
Low  &72.62 &76.62 &\underline{62.27} &73.37 &\textbf{60.57} &78.79 &70.64 & 10.3\%\\
\hline
\end{tabular}
\label{tab:high-low}
\end{table*}

\subsection{Performance Comparison}

We compare the proposed DGM-DTE model with existing OD DTE models and imbalanced regression models in terms of imbalanced DTE performance. The results are shown in Table~\ref{tab:performance}. 
For a fair comparison with imbalanced regression models~\cite{YangZCWK21,RenZY022,BrancoTR17} that usually evaluated on balanced data, we also constructed a balanced test data for performance evaluation.
We have the following findings by analyzing the experimental results: 
1) Our DGM-DTE model outperforms existing models significantly on all datasets and evaluation metrics. 
The main reason is that our model considers the data imbalance and focuses on high-shot and rare labeled data, improving the performance on rare labeled data while maintaining high-shot data performance.
For D1 dataset, the MAE of DGM-DTE outperforms existing models by 14\% - 32\%, and the MAE performance of our model improves by 17\% - 42\% on D2. For the balanced test, our model also outperforms existing imbalanced regression models, demonstrating that our model can be effectively used for imbalanced data prediction.
2) For OD DTE models, the performance of TEMP varies with the dataset, performing better on D2, because data in D1 is more substantial imbalanced than that in D2. Besides, the performance of XGBoost, STNN, and xDeepFM are unsatisfactory, indicating that only using order features is insufficient for DTE tasks. BGE achieves better performance since BGE introduces unobservable attributes of orders.
3) Some imbalanced regression models, such as LDS and SMOGN, perform well in the balanced test, but perform poorly on imbalanced real datasets. One plausible explanation is that such models focus on the rare labeled data and improve the predictive performance of rare labeled data, but at the expense of compromising the performance of high-shot data.
4) In general, D2 has better prediction performance than D1, especially on MAE and EW. The main reason is D1 suffers from severely imbalanced data, as shown in Fig.~\ref{fig:dis}, D1 has a longer tail and a large number of data in the tails.


To better understand the effect of different methods on imbalanced data, we compare the performance of different methods on high-, medium- and low-shot data, which is displayed in Table~\ref{tab:high-low}. Our DGM-DTE outperforms the OD DTE models (\ie XGBoost and BGE) on all shots. 
The imbalanced regression models, LDS and SMOGN, achieved excellent performance in low-shot data, but inferior performance in high-shot, which suggests that these models enhance the performance of low-shot data at the expense of high-shot prediction.
The im-reg is a variant of DGM-DTE, which directly uses imbalanced data as input of the dual graph module. 
The improvement shows that we can effectively improve the performance of low-shot data while ensuring high-shot performance by multi-task learning with a dual graph module for the head and tail data separately.

\begin{figure}[t]
 \centering
 \subfigure[Effectiveness of multitask learning]{
  \begin{minipage}{0.45\columnwidth}
   \centering
   \includegraphics[width=\textwidth]{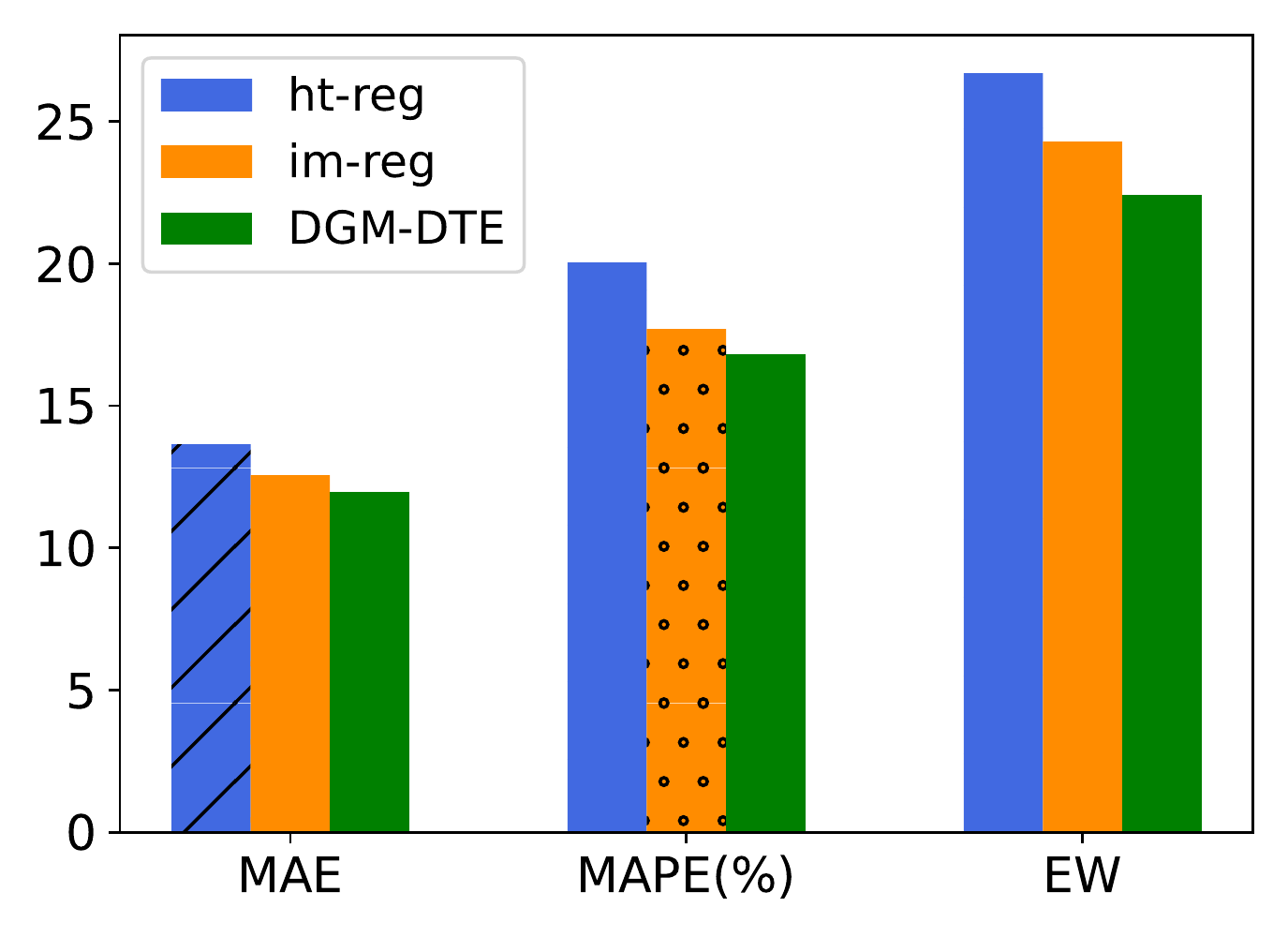}
  \end{minipage}
 }
 \subfigure[Effectiveness of dual-graph model]{
  \begin{minipage}{0.45\columnwidth}
   \centering
   \includegraphics[width=\textwidth]{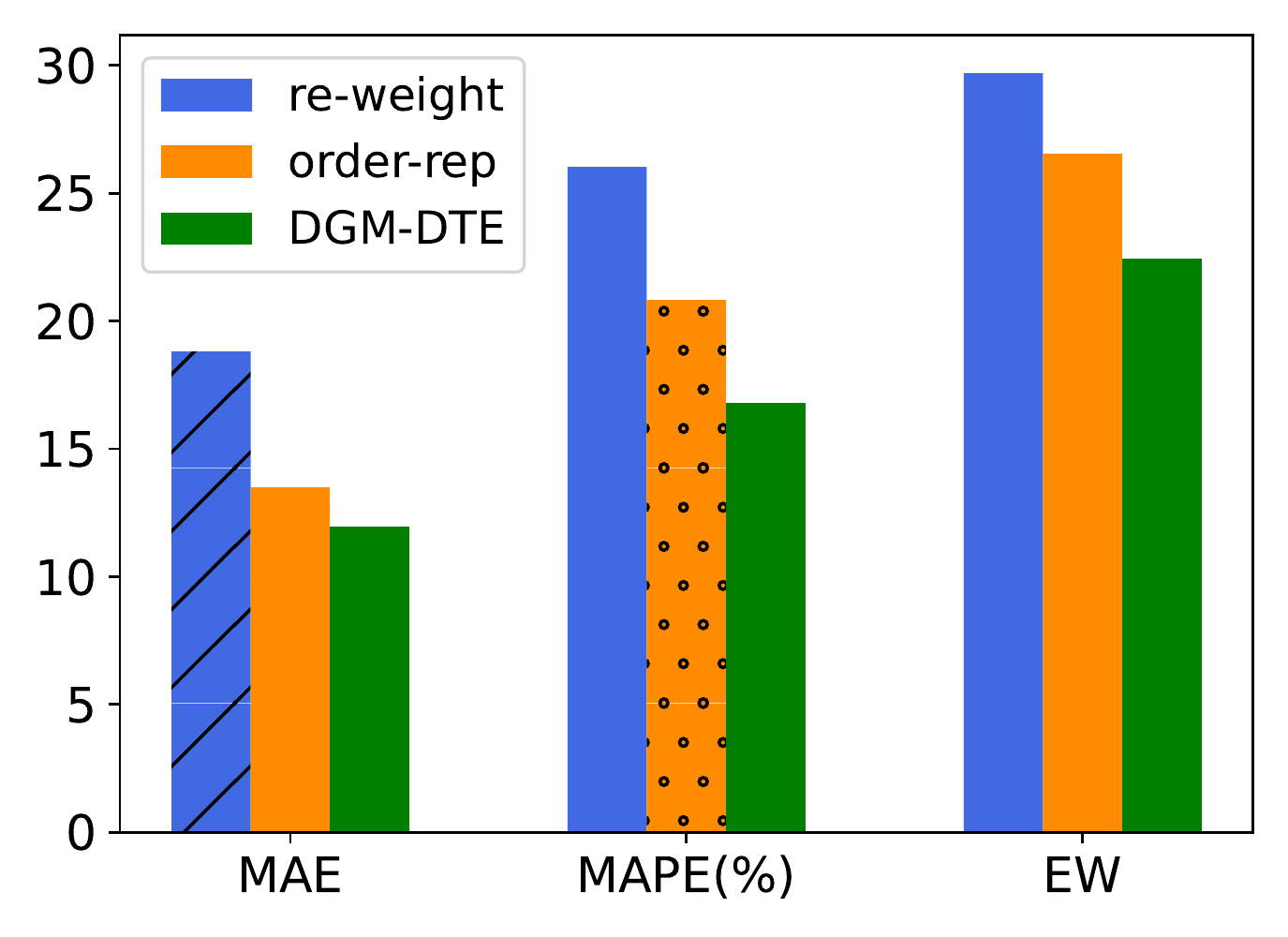}
  \end{minipage}
 }
 \caption{Ablation study of different parts of DGM-DTE model.}
 \label{fig:ablation}
\end{figure}

\subsection{Ablation Study}
We design several variants and compare their prediction performance to verify the effectiveness of different components of DGM-DTE, as shown in Fig.~\ref{fig:ablation}. 

To evaluate the effectiveness of multitask learning, we design two variants named ht-reg and im-reg. Both variants perform only regression task, where the input of ht-reg is manually divided into head and tail data, and im-reg with all imbalanced data as input for the dual graph model. 
The performance of ht-reg is worst since we cannot know whether the data is head or tail data in the test. Besides, im-reg model does not achieve better performance because it is insufficient to use re-weight to enable the model to focus more on low-shot data.

To investigate the effectiveness of the dual graph module, we design two variants: order representation model (short as order-rep, which only uses head data learning module with all data as input), and feature re-weight model (short as re-weight, which only uses tail data learning module with all data as input). The re-weight is a kind of imbalanced regression model, which pays more attention to rare labeled data, leading to poor performance on real imbalanced datasets. The order-rep does not consider the imbalance of the data and only uses GNN for order representation, which also fails to achieve satisfactory results.

\subsection{Parameter Analysis}
\begin{figure}[t]
 \centering
 \subfigure[Order embedding size]{
  \begin{minipage}{0.4\columnwidth}
   \centering
   \includegraphics[width=\textwidth]{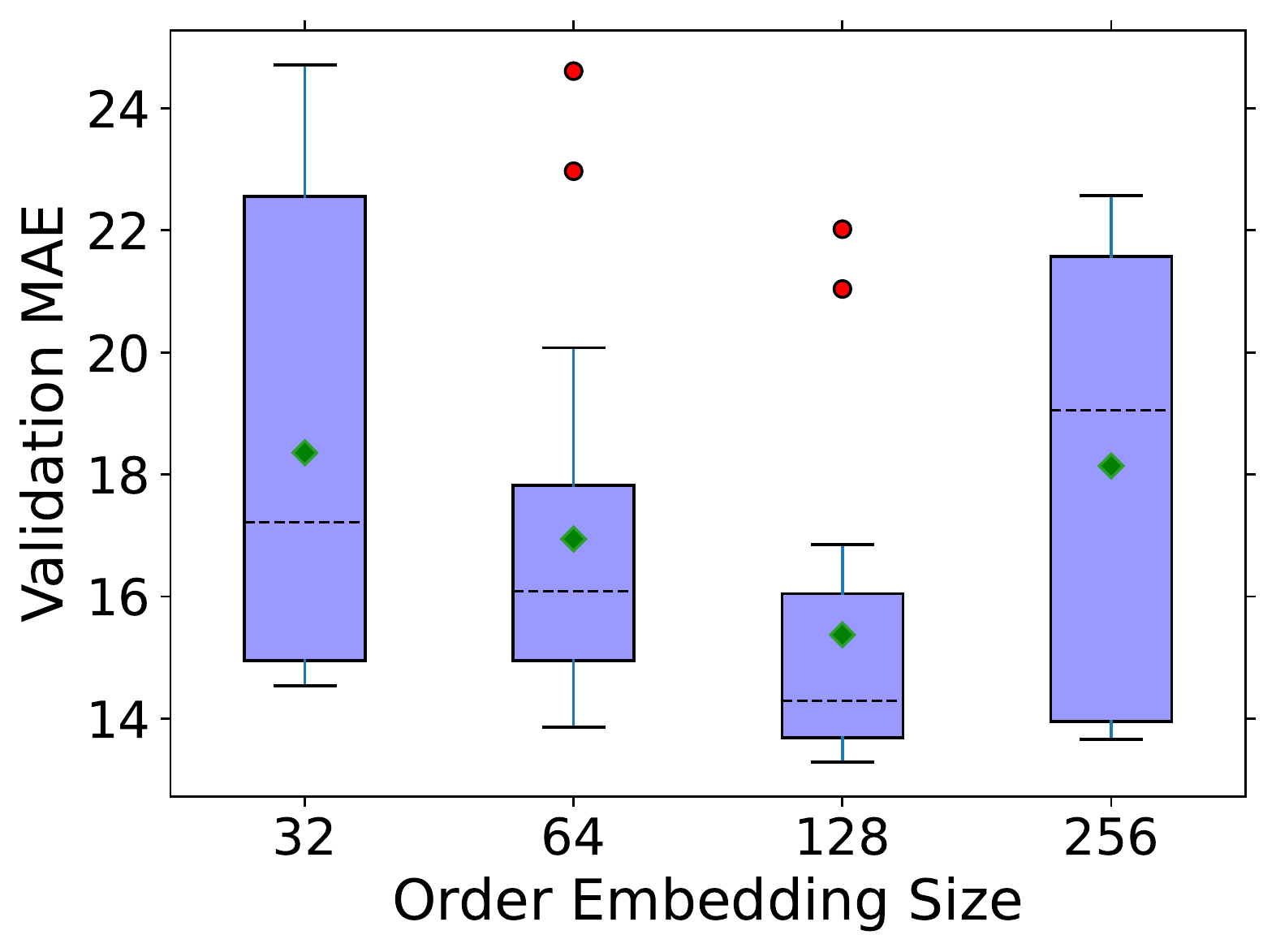}
  \end{minipage}
 }
 \subfigure[Classification thresholds]{
  \begin{minipage}{0.4\columnwidth}
   \centering
   \includegraphics[width=\textwidth]{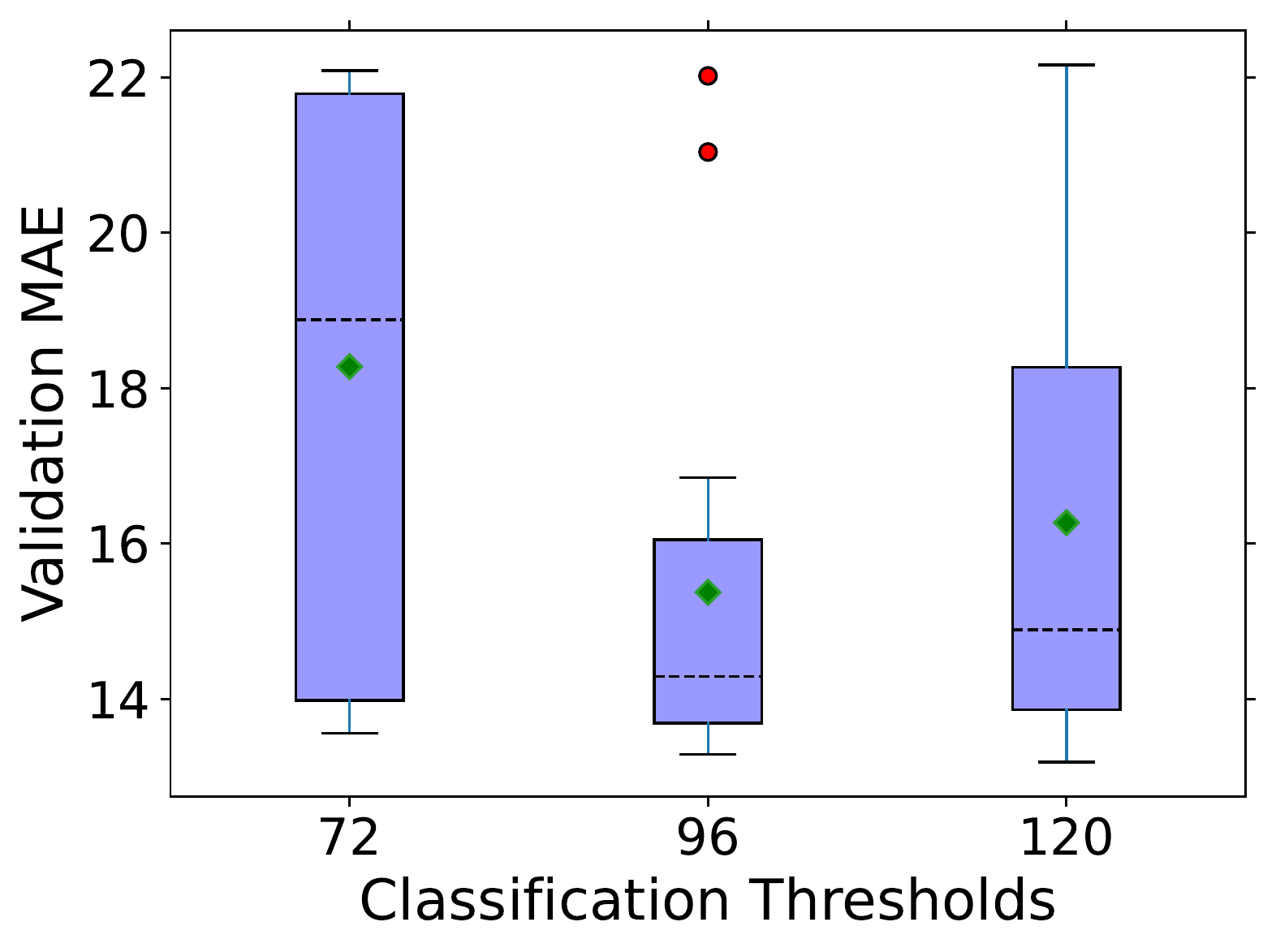}
  \end{minipage}
 }
 \caption{Performance of DGM-DTE with different parameters.}
 \label{fig:para}
\end{figure}

We analyze the sensitivity of the important parameters for prediction performance. The main parameters of DGM-DTE are order embedding size $d_{O}$ and classification threshold $t_c$. To better observe the effect of different parameters on training, we analyze the MAE on the validation set, as shown in Fig.~\ref{fig:para}. For order embedding size, the MAE decreases and then increases significantly as $d_{O}$ becomes larger, and shows optimal MAE performance at the size of 128. The MAE performs optimally at a classification threshold of 96 hours, while the predictive performance is unstable at values of 72 and 120 hours.

\section{Conclusion}
This paper presents a novel dual graph multitask framework for e-commerce delivery time estimation, which addresses the prevalent data imbalanced problem in the industry. We first classify the data into head and tail data depending on the delivery time, and then use a dual graph-based representation module to separately deal with the head and tail data from the classification, enabling the model to focus on both the high-shot data in the head and the rare labeled data in the tail. Finally, we aggregate two parts of data and estimate the delivery time. Experimental results on real datasets show that DGM-DTE can effectively improve the overall prediction performance, while also improving the predictive capability of rare labeled orders.

\section*{Acknowledgements}
This work is partially supported by NSFC No.62202279; National Key R\&D Program of China No. 2021YFF0900800; Shandong Provincial Key Research and Development Program (Major Scientific and Technological Innovation Project) (No. 2021CXGC010108); Shandong Provincial Natural Science Foundation (No. ZR202111180007); the Fundamental Research Funds of Shandong University. This work is also supported, in part, by Alibaba Group through the Alibaba Innovative Research (AIR) Program and the Alibaba-NTU Singapore Joint Research Institute (AN-GC-2021-008-02); 
the State Scholarship Fund by the China Scholarship Council (CSC).
%
%
%
%


\end{document}